\documentclass[10pt,twocolumn,letterpaper]{article}

\usepackage{wacv}
\usepackage{times}
\usepackage{epsfig}
\usepackage{graphicx}
\usepackage{amsmath}
\usepackage{amssymb}
\usepackage{url}
\usepackage{multirow}
\usepackage{caption,subcaption}
\usepackage{enumitem}
\usepackage[super]{nth}
\usepackage[ruled,linesnumbered]{algorithm2e}
\usepackage{setspace}
\usepackage{pifont}

\makeatletter
\newcommand{\appendixhead}%
{\centering {\huge Supplementary}
\vspace{0.25in}}
\makeatother

\newcommand{\cmark}{\ding{51}}%
\newcommand{\xmark}{\ding{55}}%

\wacvfinalcopy 

\ifwacvfinal
\fi

\ifwacvfinal
\usepackage[breaklinks=true,bookmarks=false]{hyperref}
\else
\usepackage[pagebackref=true,breaklinks=true,colorlinks,bookmarks=false]{hyperref}
\fi

\begin{document}

\title{Data InStance Prior (DISP) in Generative Adversarial Networks}

\author{Puneet Mangla\thanks{Authors contributed equally   } $^{1}$ \\
{\tt\small pmangla@adobe.com}
\and
Nupur Kumari$^{*2}$ \thanks{Work done while at Adobe} \\
{\tt\small nupurkmr9@gmail.com}
\and
Mayank Singh$^{*2 \ \dagger}$  \\
{\tt\small mayanksingh027@gmail.com}
\and
Balaji Krishnamurthy$^{1}$ \\
{\tt\small kbalaji@adobe.com}\\
\and
Vineeth N Balasubramanian$^{3}$ \\
{\tt\small vineethnb@iith.ac.in}
\and
$^1$Media and Data Science Research lab, Adobe \quad $^2$CMU \quad $^3$IIT Hyderabad, India
}

\maketitle

\begin{abstract}
Recent advances in generative adversarial networks (GANs) have shown remarkable progress in generating high-quality images. However, this gain in performance depends on the availability of a large amount of training data. In limited data regimes, training typically diverges, and therefore the generated samples are of low quality and lack diversity. Previous works have addressed training in low data setting by leveraging transfer learning and data augmentation techniques. We propose a novel transfer learning method for GANs in the limited data domain by leveraging informative data prior derived from self-supervised/supervised pre-trained networks trained on a diverse source domain. We perform experiments on several standard vision datasets using various GAN architectures (BigGAN, SNGAN, StyleGAN2) to demonstrate that the proposed method effectively transfers knowledge to domains with few target images, outperforming existing state-of-the-art techniques in terms of image quality and diversity. We also show the utility of data instance prior in large-scale unconditional image generation.
\end{abstract}

\section{Introduction}
Generative Adversarial Networks (GANs) are at the forefront of modern high-quality image synthesis in recent years \cite{biggan,stylegan2,ffhq_stylegan}. GANs have also demonstrated excellent performance on many related computer vision tasks such as image manipulation \cite{imgmani2017CycleGAN,imgmanipix2pix2017}, image editing \cite{controlling2020iclr,interface2020shen,gansteerability} and compression \cite{compress2018gan}. Despite the success in large-scale image synthesis, GAN training suffers from a number of drawbacks that arise in practice, such as training instability and mode collapse \cite{goodfellow2016deep,arora2017generalization}. This has prompted research in several non-adversarial generative models \cite{glann2019jitendra,glo,imle,vae2_2013kingma}. These techniques are implicitly designed to overcome the mode collapse problem, however, the quality of generated samples are still not on par with GANs.

Current state-of-the-art deep generative models require a large volume of data and computation resources. The collection of large datasets of images suitable for training - especially labeled data in case of conditional GANs - can also be a daunting task due to issues such as copyright, image quality. To curb these limitations, researchers have recently proposed techniques inspired by transfer learning \cite{bsa2019harada,transfergan2018wang,freezeD2020jinwoo} and data augmentation \cite{aug2020karras,diffaug2020zhao,crgan2019consistency}. Advancements in data and computation efficiency for image synthesis can enable its applications in data-deficient fields such as medicine \cite{medicine2019generative} where labeled data procurement can be difficult.

Transfer learning is a promising area of research \cite{transfer2_2014_CVPR,transfer4_2009survey} that leverages prior information acquired from large datasets to help in training models on a target dataset under limited data and resource constraints. There has been extensive exploration of transfer learning in classification problems that have shown excellent performance on various downstream data-deficient domains. Similar extensions of reusing pre-trained networks for transfer learning (i.e. fine-tuning a subset of pre-trained network weights from a data-rich domain) have also been recently employed for image synthesis in GANs \cite{transfergan2018wang,bsa2019harada,freezeD2020jinwoo,minegan_2020_CVPR,pretraingan2020carin} in the limited data regime. However, these approaches are still prone to overfitting on the sparse target data, and hence suffer from degraded image quality and diversity.

In this work, we propose a simple yet effective way of transferring prior knowledge in unsupervised image generation given a small sample size ($\sim 100$-$2000$) of the target data distribution. Our approach is motivated by the formulation of the IMLE technique \cite{imle} that seeks to obtain mode coverage of target data distribution by learning a mapping between latent and target distributions using a maximum likelihood criterion. We instead propose the use of data priors in GANs to match the representation of the generated samples to real modes of data. In particular, we show that using an informative \textit{data instance prior} in limited and large-scale unsupervised image generation substantially improves the performance of image synthesis. We show that these data priors can be derived from commonly used computer vision pre-trained networks \cite{vgg16,zhang2018perceptual,bsa2019harada,glann2019jitendra} or self-supervised data representations \cite{simclr} (without any violation of the target setting's requirements, i.e. ensuring that the pre-trained network has not been trained on few-shot classes in the few-shot learning setting, for instance). In case of sparse training data, our approach of using data instance priors leverages a model pre-trained on a rich source domain to learn the target distribution. Different from previous works \cite{bsa2019harada, minegan_2020_CVPR, transfergan2018wang} which rely on fine-tuning models trained on a data-rich domain, we propose to leverage the feature representations of the source model as data instance priors, to distill knowledge \cite{distill2015bengio,distill2015hinton} into the target generative problem setting.

We note that our technique of using instance level priors for transfer learning becomes fully unsupervised in case the data priors are extracted from self-supervised pre-trained networks. Furthermore, in addition to image generation in low data domain, we also achieve state-of-the-art  Fr\'echet inception distance (FID) score \cite{fidmetric} on large-scale unsupervised image generation.

We summarize our main contributions as follows:
\begin{itemize}[noitemsep,topsep=0pt]
\item We propose Data InStance Prior (DISP), a novel transfer learning technique for GAN image synthesis in low-data regime. We show that employing DISP in conjunction with existing few-shot image generation methods outperforms state-of-the-art results. We show with as little as $100$ images our approach DISP results in generation of diverse and high quality images (see Figure \ref{fig:few_shot_example_more}). 
\item We demonstrate the utility of our approach in large-scale unsupervised GANs \cite{sngan, biggan}. It achieves the new state-of-the-art in terms of image quality \cite{fidmetric} and diversity \cite{precision_recall_distributions, ivom2017metz}.
\end{itemize}

We call our method a \textit{data instance prior} (and not just data prior), since it uses representations of instances as a prior, and not a data distribution itself.

\section{Related Work}
\paragraph{Deep Generative Models} In recent years, there has been a surge in the research of deep generative models. Some of the popular approaches include variational auto-encoders (VAEs) \cite{vae1_2014stochastic,vae2_2013kingma}, auto-regressive (AR) models \cite{ar1_2016pixel,ar2_2016conditional} and GANs \cite{gan_goodfellow}. VAE models learn by maximizing the variational lower bound of training data likelihood. Auto-regressive approaches model the data distribution as a product of the conditional probabilities to sequentially generate data. GANs comprise of two networks, a generator and a discriminator that train in a min-max optimization. Specifically, the generator aims to generate samples to fool the discriminator, while the discriminator learns to distinguish these generated samples from the real samples. Several research efforts in GANs have focused on improving the performance \cite{pggan2018celeba,denton2015deep,radford2015unsupervised,stylegan2,ffhq_stylegan,biggan,crgan2019consistency} and training stability \cite{salimans2016improved,gulrajani2017improved,arjovsky2017wasserstein,sngan,mao2017least, ssgan}. Recently, the areas of latent space manipulation for semantic editing \cite{interface2020shen,gansteerability,zhu2020indomain,controlling2020iclr} and few-shot image generation \cite{minegan_2020_CVPR,freezeD2020jinwoo,bsa2019harada} have gained traction in an effort to mitigate the practical challenges while deploying GANs. Several other non-adversarial training approaches such as \cite{glann2019jitendra,glo,imle,yu2020inclusive} have also been explored for generative modeling, which leverage supervised learning along with perceptual loss \cite{zhang2018perceptual} for training such models.

\vspace{-8pt}
\paragraph{Transfer Learning in GANs} While there has been extensive research in the area of transfer learning for classification models \cite{transfer2014benjio,transfer2_2014_CVPR,transfer3_2015simultaneous,transfer4_2009survey,transfer5_2014decaf}, relatively fewer efforts have explored this on the task of data generation \cite{transfergan2018wang,minegan_2020_CVPR,bsa2019harada,pretraingan2020carin,freezeD2020jinwoo}. \cite{transfergan2018wang} proposed to fine-tune a pre-trained GAN model (often having millions of parameters) from a data-rich source to adapt to the target domain with limited samples. This approach, however, often suffers from overfitting as the final model parameters are updated using only few samples of the target domain. To counter overfitting, the work of \cite{bsa2019harada} proposes to update only the batch normalization parameters of the pre-trained GAN model. In this approach, however, the generator is not adversarially trained and uses supervised $L_1$ pixel distance loss and perceptual loss \cite{johnson2016perceptual,zhang2018perceptual} which often leads to generation of blurry images in the target domain. Based on the assumption that source and target domain support sets are similar, \cite{minegan_2020_CVPR} recently proposed to learn an additional mapping network that transforms the latent code suitable for generating images of target domain while keeping the other parameters frozen. We show that our method DISP outperforms the leading baselines in few-shot image generation including \cite{bsa2019harada, minegan_2020_CVPR, diffaug2020zhao}.

\begin{figure*}[t]
\centering
\scalebox{0.8}{
    \includegraphics[width=\textwidth]{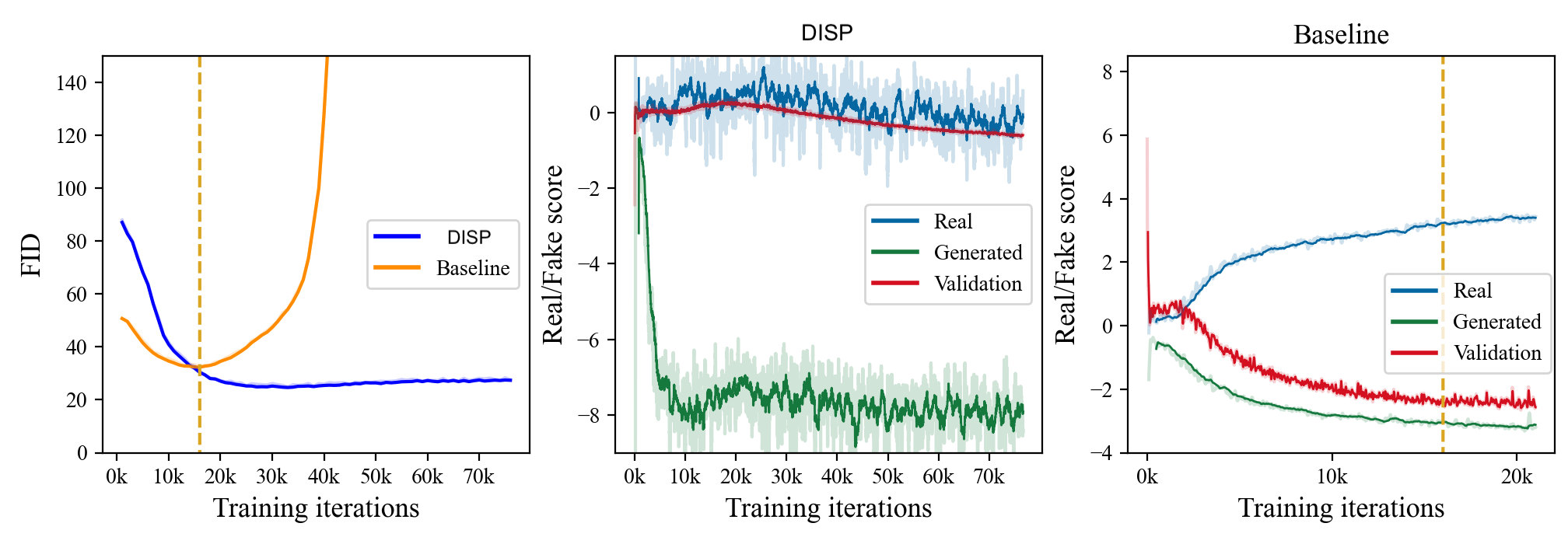}}
    \caption{\footnotesize{Comparison between DISP and Baseline when trained on $10\%$ data of CIFAR-100. \textit{left}: FID on training set for baseline training starts to increase early (around 15k iterations) unlike the FID curve of DISP training. \textit{middle}:  Discriminator score on training and validation images remain similar to each other and consistently higher than score of generated images for DISP model.  \textit{right}:  Discriminator score on training and validation images diverges and the training collapses for the baseline model.}}
    \label{fig:discriminator_overfitting}
\end{figure*}

A related line of recent research aims to improve large-scale unsupervised image generation in GANs by employing self-supervision - in particular, an auxiliary task of rotation prediction \cite{ssgan} or using one-hot labels obtained by clustering in the discriminator’s \cite{liu2020selfconditioned} or ImageNet classifier feature space \cite{sage2018logo}. In contrast, our method utilizes instance level priors derived from the feature activations of self-supervised/supervised pre-trained networks to improve unconditional few-shot and large-scale image generation, leading to simpler formulation and better performance as shown in our experiments in Section \ref{sec:large_scale} and Supplementary. Recently, some methods \cite{aug2020karras,diffaug2020zhao,crgan2019consistency,icrgan2020improved} have leveraged data augmentation to effectively increase the number of samples and prevent overfitting in GAN training. However, data augmentation techniques often times alter the true data distribution and there is a leakage of these augmentations to the generated image, as shown in \cite{icrgan2020improved,diffaug2020zhao}. To overcome this, \cite{diffaug2020zhao} recently proposed to use differential augmentation and \cite{aug2020karras} leveraged an adaptive discriminator augmentation mechanism. We instead focus on leveraging informative instance level priors and also show how our method can be used in conjunction with augmentation techniques \cite{diffaug2020zhao} to further improve the performance.

\section{Preliminaries}\label{background}
We briefly describe Conditional Generative Networks(cGANs) before discussing our methodology. cGANs consists of a generator network $G$ which is trained adversarially with a discriminator network $D$ to learn a target data distribution $q(\mathbf{x}\vert{y})$. Given a noise vector $z$ and a condition vector $y$, $G$ generates a sample $x$ e.g. an image and the role of $D$ is to distinguish between real samples and those generated from $G$. Conditional GANs use auxiliary information $y$ for e.g. class label of the sample as input in the generator and discriminator networks. The standard hinge loss \cite{hinge_loss_gan} for training cGANs is given by:
\begin{equation}
\begin{aligned}
   & \begin{aligned}
       L_{D} & =  {\mathbb{E}_{y \sim {q(y)}}} \big[ {\mathbb{E}_{\mathbf{x} \sim q(\mathbf{x}\vert y)}}[\max(0,1- D(\mathbf{x},y))] \big] \\
       & + {\mathbb{E}_{y \sim {q(y)}}} \big[  {\mathbb{E}_{\mathbf{z} \sim p(\mathbf{z})}} [\max(0, 1 + D(G(\mathbf{z}\vert y),y))]  \big]
   \end{aligned} \\ 
   & L_{G} = - {\mathbb{E}_{y \sim {q(y)}}} \big[ {\mathbb{E}_{\mathbf{z} \sim p(\mathbf{z})}} [  D(G(\mathbf{z}\vert y), y) ]  \big]
\end{aligned}\label{eq:gan_standard}
\end{equation}

where the discriminator score $D(\mathbf{x},y)$ depends on input image (either real or fake) and conditional label ${y}$ \cite{cgan_projection,odena2017conditional}. The label information is generally passed into $G$ through a one-hot vector concatenated with $z$ or through conditional batch norm layers \cite{conditional_bn_1, conditional_bn_2}. 

\section{Methodology}
We propose a transfer learning framework, Data InStance Prior (DISP), for training GANs that exploits knowledge extracted from self-supervised/supervised networks, pre-trained on a rich and diverse source domain in the form of instance level priors. GANs are observed to be prone to mode-collapse that is further exacerbated in case of sparse training data. It has been shown that providing class label information in GANs significantly improves training stability and quality of generated images as compared to unconditional setting \cite{cgan_projection, ssgan}. We take motivation from the reconstructive framework of IMLE \cite{imle} and propose to condition GANs on image instance prior that acts as a regularizer to prevent mode collapse and discriminator overfitting.

\begin{figure*}[t]
\centering
\scalebox{0.65}{
    \includegraphics[width=\linewidth]{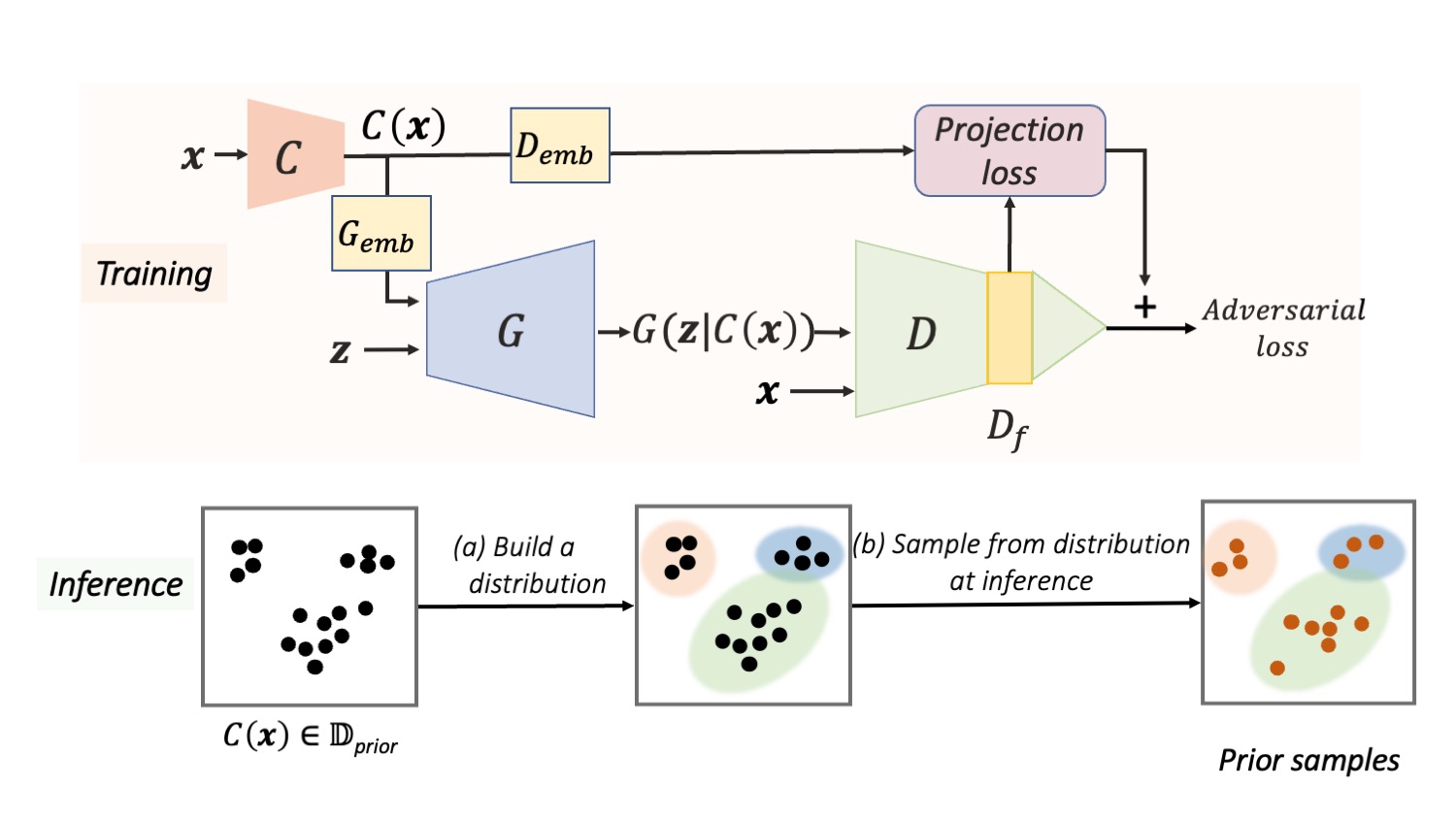}}
    
    \caption{\footnotesize{Overview of our proposed technique, Data Instance Priors (DISP) for transfer learning in GANs. \underline{\textit{Top:}} DISP training with feature $C(\mathbf{x})$ of a real sample $\mathbf{x}$ as a conditional prior in the conditional GAN framework of \cite{cgan_projection}. $C$ is a pre-trained network on a rich source domain
    from which we wish to transfer knowledge. \underline{\textit{Bottom}}: Inference over trained GAN involves learning a distribution over the set of training data prior $\{C(\mathbf{x})\}$ to enable sampling of conditional priors.}}
    \label{fig:block_diagram}
\end{figure*}

\paragraph{Knowledge Transfer in GAN}\label{our_method}
GANs are a class of implicit generative models that minimize a divergence measure between the data distribution $q(\mathbf{x})$ and the generator output distribution $G(\mathbf{z})$ where $\mathbf{z} \sim p(\mathbf{z})$ denotes the latent distribution. Intuitively, the minimization of this divergence objective ensures that each generated sample $G(\mathbf{z})$ is close to some data example $\mathbf{x} \sim q(\mathbf{x})$. However, this does not ensure the converse, i.e. each real sample has a generated sample in its vicinity, which can result in mode dropping. To counter this, especially in limited data regime, we propose to update the parameters of the model so that each real data example is close to some generated sample similar to \cite{imle} by using data instance priors as conditional label in GANs. We use the image features extracted from networks pre-trained on a large source domain as the instance level prior to enable transfer of knowledge.

Given a pre-trained feature extractor $C:\mathbb{R}^p \rightarrow \mathbb{R}^{d}$, which is trained on a source domain using supervisory signals or self-supervision, we use its output $C(\mathbf{x})$ as the conditional information during GAN training. $G$ is conditioned on $C(\mathbf{x})$ using conditional batch-norm \cite{conditional_bn_2} whose input is $G_{emb}(C(\mathbf{x}))$, where $G_{emb}$ is a learnable projection matrix. During training we enforce that $G(\mathbf{z}|C(\mathbf{x}))$ is close to the real image $\mathbf{x}$ in discriminator feature space. Let the discriminator be $D = D_l \circ D_f$ ($\circ$ denotes composition) where $D_f$ is discriminator's last feature layer and $D_l$ is the final linear classifier layer. To enforce the above objective we map $C(\mathbf{x})$ to discriminator's feature layer dimension using a trainable projection matrix $D_{emb}$ and minimize distance between $D_{emb}(C(\mathbf{x}))$ and $D_f$ of both real image $\mathbf{x}$ and generated image $G(\mathbf{z}|C(\mathbf{x}))$ in an adversarial manner. Hence, our final GAN training loss for the discriminator and generator is given by: 

\begin{equation}
\begin{aligned}
    L_{D} = & 
     {\mathbb{E}_{ \mathbf{{x}} \sim q(x)}}
    [\max(0, 1 - D(\mathbf{{x}} , C(\mathbf{{x}}))) ] \\
    & + {\mathbb{E}_{ \mathbf{{x}} \sim q(x), \mathbf{z} \sim p(\mathbf{z})}}  
    [\max(0, 1 + D(G(\mathbf{z}|C(\mathbf{{x}})) , C(\mathbf{{x}}) ))  ]   \\
   L_{G} = & - {\mathbb{E}_{ \mathbf{x} \sim q(x) ,\mathbf{z} \sim p(\mathbf{z}) }} [ D(G(\mathbf{z}|C(\mathbf{x})) , C(\mathbf{x}))]
\end{aligned}\label{eq:feature_match}
\end{equation}
where
\begin{equation}
\begin{aligned}   
 D(\mathbf{x},\mathbf{y}) & = D_{emb}(\mathbf{y})\boldsymbol{\cdot}  D_f(\mathbf{{x}}) + D_l \circ D_f(\mathbf{{x}})
\end{aligned}\label{eq:feature_match2}
\end{equation}
In the above formulation, the first term in Eq. \ref{eq:feature_match2} is the projection loss as in \cite{cgan_projection} between input image and conditional embedding of discriminator. Since conditional embedding is extracted from a pre-trained network, above training objective leads to feature level knowledge distillation from $C$. It also acts as a regularizer on the discriminator reducing its overfitting in the limited data setting. As shown in Figure \ref{fig:discriminator_overfitting}, the gap between discriminator score ($D_l \circ D_f$) of training and validation images keeps on increasing and FID quickly degrades for baseline model as compared to DISP when trained on only $10\%$ data of CIFAR-100. Moreover, enforcing feature $D_f(G(\mathbf{z}|C(\mathbf{x})))$ to be similar to $D_{emb}(C(\mathbf{x}))$ promotes that for each real sample, there exists a generated sample close to it and hence promotes mode coverage of target data distribution. We demonstrate that the above proposed use of data instance priors from a pre-trained feature extractor, while designed for a limited data setting, also benefits in large-scale image generation. Our overall methodology is illustrated in Figure \ref{fig:block_diagram} and pseudo code is provided in the Supplementary section.

\paragraph{Random image generation at inference}  Given the training set $\mathbb{D}_{image}=
\{{\mathbf{x}_j}\}_{j=1}^{n}$ of sample size $n$ and its corresponding training data prior set $\mathbb{D}_{prior}= \{{C(\mathbf{x}_j})\}_{j=1}^{n}$, the generator requires access to $\mathbb{D}_{prior}$ for sample generation. In case of few-shot and limited image generation where size of $\mathbb{D}_{prior}$ is less, to create more variations, we generate images conditioned on prior samples from a vicinal mix distribution i.e
\begin{equation}
\begin{aligned}
& G(\mathbf{z} | \mathbf{p}) & \textnormal{where}\; \mathbf{p} \sim \mathcal{V}_{mix}
\end{aligned}
\end{equation}
The vicinal mix distribution is defined as:
\begin{equation}
\begin{aligned}
& \mathcal{V}_{mix}(\mathbf{p}) = \frac{1}{\vert \mathbb{D}_{prior} \vert^2}\sum_{i, j}^{\vert \mathbb{D}_{prior} \vert } \mathbb{E}_\lambda \left [ \delta(\lambda \cdot \mathbf{p_i} + (1-\lambda) \cdot \mathbf{p_j}) \right ] \\ & \textnormal{where} \; \; \lambda \sim \mathcal{U}[0,1] \; \; \textnormal{and} \; \; \delta(.) \; \; \textnormal{is dirac-delta function}
\end{aligned}
\end{equation}

In case of large-scale image generation,
we learn a Gaussian Mixture Model (GMM) \cite{gmm} on $\mathbb{D}_{prior}$. This enables memory efficient sampling of conditional prior from the learned GMM distribution during inference:
\begin{equation}
    \begin{aligned}
     G(\mathbf{z} | \mathcal{N}(\mathbf{\mu},\mathbf{\Sigma} )) \;\; \textnormal{where}\; \mathbf{\mu} , \mathbf{\Sigma} \sim \textnormal{GMM}( G_{emb}(\mathbb{D}_{prior}))
    \end{aligned} 
\end{equation}

\section{Experiments}
\label{sec_expts}
We perform extensive experiments to highlight the efficacy of our data instance prior module, DISP in unsupervised training based on SNGAN \cite{sngan}, BigGAN \cite{biggan} and StyleGAN2 \cite{stylegan2} architectures. For extracting image prior information, we use the last layer activations of: Vgg16 \cite{vgg16} classification network trained on ImageNet; and SimCLR \cite{simclr} network trained using self-supervision on ImageNet. We conduct experiments on (1) \textit{few-shot} ($\sim25$-$100$ images), (2) \textit{limited} ($\sim2$k-$5$k images) and (3) \textit{large-scale} ($\sim50$k-$1$M images) data settings. For evaluation, we use FID \cite{fidmetric}, precision and recall scores \cite{kynkaanniemi2019improved_pr} to assess the quality and mode-coverage/diversity of the generated images.

\begin{table}[t]
\scalebox{0.7}{
    \begin{tabular}{l c r r r r r r}
    \hline
    & & \multicolumn{6}{c}{SNGAN (128 x 128)}   \\
    \textbf{Method} & \begin{tabular}{@{}c@{}}\textbf{Pre-} \\ \textbf{training}\end{tabular} &  \multicolumn{3}{c}{\textbf{Anime}} &  \multicolumn{3}{c}{\textbf{Faces}} \\
    &  & FID $\downarrow$ & P $\uparrow$ & R $\uparrow$ & FID $\downarrow$ & P $\uparrow$ & R $\uparrow$ 
    \\  \hline &    &  &  & &  &  &\\
    From scratch &  \xmark & 120.38 &  0.61 & 0.00 &  140.66 & 0.31 & 0.00 \\
    + DISP-Vgg16 & & \textbf{66.85}  &\textbf{0.71} & \textbf{0.03} & \textbf{68.49} & \textbf{0.74}  & \textbf{0.15} \\
    \hline &  & &   &  & &  &  \\
    TransferGAN & \cmark  &102.75 & \textbf{0.70} & 0.00 &  101.15  & \textbf{0.85} & 0.00 \\
     + DISP-Vgg16 & &\textbf{86.96} & 0.57 & \textbf{0.02} & \textbf{75.21} & 0.70 & \textbf{0.10} \\
    \hline &  & &   &  & &  &  \\
    FreezeD & \cmark &109.40  & \textbf{0.67} & 0.00 & 107.83 & \textbf{0.83} & 0.00   \\
    + DISP-Vgg16 & & 93.36 & 0.56  & \textbf{0.03} & 77.09  & 0.68 & 0.14 \\
    + DISP-SimCLR & & \textbf{89.39} & 0.46  & 0.025 & \textbf{70.40}  & 0.74 & \textbf{0.22} \\
    \hline & &  &    & & &  &  \\
    ADA & \xmark & 78.28  & 0.87  & 0.0 &  159.3 & 0.69  & 0.0 \\
    + DISP-Vgg16 & & \textbf{60.8}   & \textbf{0.90}  & \textbf{0.003} &  \textbf{79.5}  & \textbf{0.85} & \textbf{0.004} \\
    \hline & &  &    & & &  &  \\
    DiffAugment & \xmark & 85.16  & \textbf{0.95}  & 0.00 &  109.25 & \textbf{0.84}  & 0.00\\
    + DISP-Vgg16 & & \textbf{48.67}   & 0.82 & 0.03 &  \textbf{62.44}  & 0.80 & 0.19 \\
    + DISP-SimCLR & & 52.41   & 0.77 & \textbf{0.04} &  64.53  & 0.78 & \textbf{0.22}\\
    \hline &    &   & & &  & & \\
    BSA* & \cmark & 92.0  & - & - & 123.2  & - & - \\
    \begin{small}GLANN + DISP-Vgg16\end{small} & & \textbf{67.07}  & \textbf{0.87} & \textbf{0.01} & \textbf{60.11}  & \textbf{0.95} & \textbf{0.08} \\
    \hline
    \end{tabular}}
    \caption{\footnotesize{Few-shot image generation performance using 100 training images ($\downarrow$: lower is better; $\uparrow$: higher is better). Precision and Recall scores are based on \cite{kynkaanniemi2019improved_pr}. FID is computed using $10$k, $7$k generated and $10$k, $7$k real samples for Anime and Faces respectively. * denotes directly reported from the paper.}}
\end{table}\label{table:fewshot}
 
\subsection{Few-Shot Image Generation}\label{sec:few_shot}

\paragraph{\textbf{Baselines and Datasets}} We compare and augment our methodology DISP with training SNGAN from scratch and the following leading baselines: Batch Statistics Adaptation (BSA) \cite{bsa2019harada}, TransferGAN \cite{transfergan2018wang}, FreezeD \cite{freezeD2020jinwoo}, ADA \cite{ada} and DiffAugment \cite{diffaug2020zhao}. In case of BSA, a non-adversarial variant, GLANN \cite{glann2019jitendra} is used which optimizes for image embeddings and generative model through perceptual loss\footnote{The code provided with BSA was not reproducible, and hence this choice}. We use our data priors to distill knowledge over these image embeddings. For more training and hyperparameter details, please refer to Supplementary. 

We perform experiments on randomly chosen $100$ images at $128\times128$ resolution from:  (1) Anime\footnote{www.gwern.net/Danbooru2018} and (2) FFHQ \cite{ffhq_stylegan} datasets. The above choice of datasets follows from the prior work BSA. For methods with pre-training, we finetune SNGAN pre-trained on ImageNet as done in \cite{bsa2019harada} (there is no class label intersection of the above datasets with ImageNet classes). We also show additional results at $256\times256$ resolution on additional datasets (Pandas, Grumpy Cat, Obama) with StyleGAN2 \cite{stylegan2} in supplementary.

\begin{figure*}[t]
\begin{minipage}{1.\textwidth}
\begin{subfigure}{0.33\textwidth}
  \includegraphics[width=\linewidth]{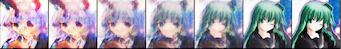}
  \includegraphics[width=\linewidth]{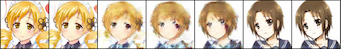}
  \includegraphics[width=\linewidth]{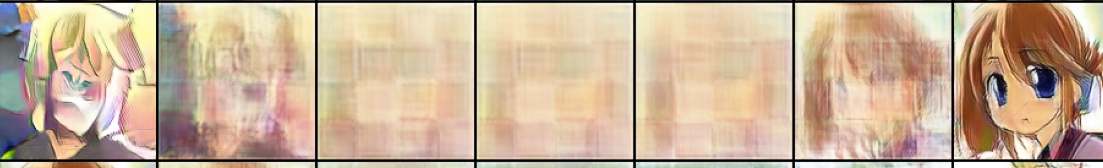}
  \includegraphics[width=\linewidth]{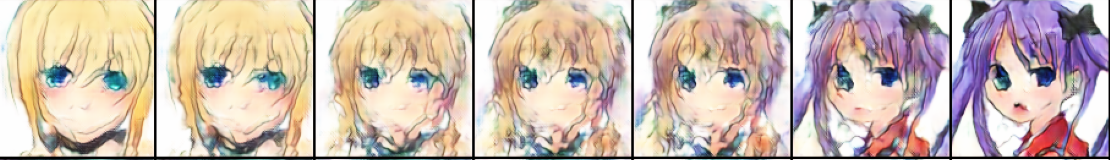}
  \includegraphics[width=\linewidth]{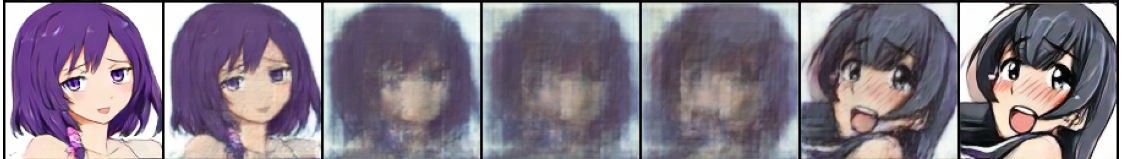}
  \includegraphics[width=\linewidth]{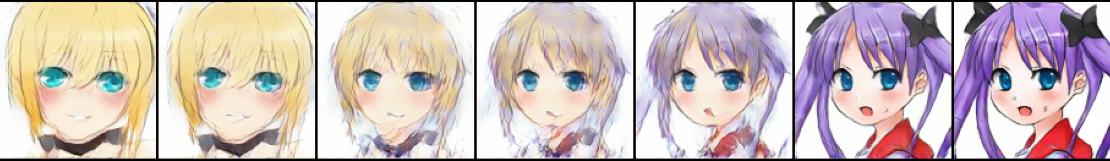}
  \caption{}
  \label{fig:2}
\end{subfigure} \hfil 
\begin{subfigure}{0.33\textwidth}
    \includegraphics[width=\linewidth]{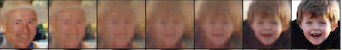}
  \includegraphics[width=\linewidth]{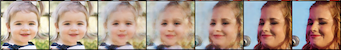}
  \includegraphics[width=\linewidth]{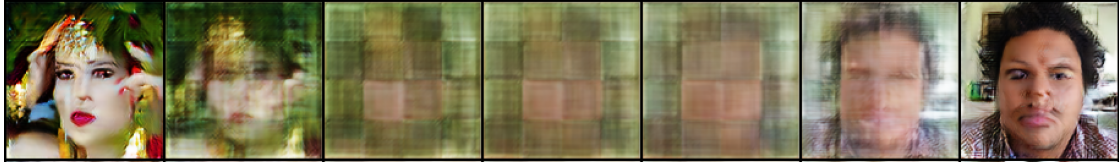}
  \includegraphics[width=\linewidth]{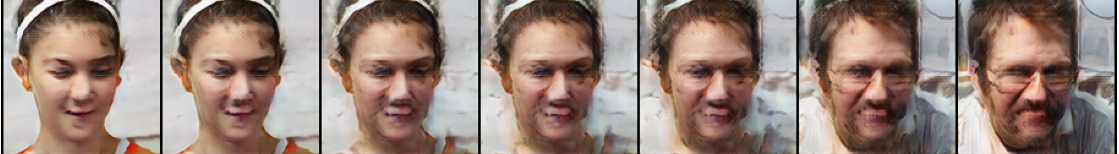}
  \includegraphics[width=\linewidth]{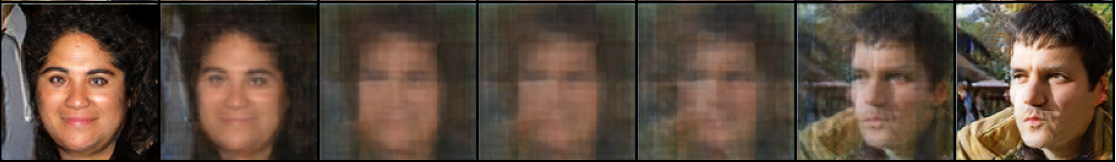}
  \includegraphics[width=\linewidth]{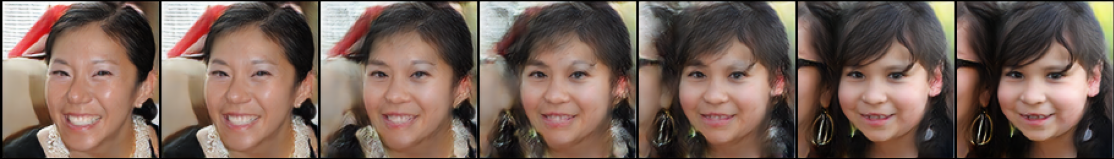}
  \caption{}
  \label{fig:2}
\end{subfigure} \hfil 
\begin{subfigure}{0.33\textwidth}
  \includegraphics[width=1.1\linewidth]{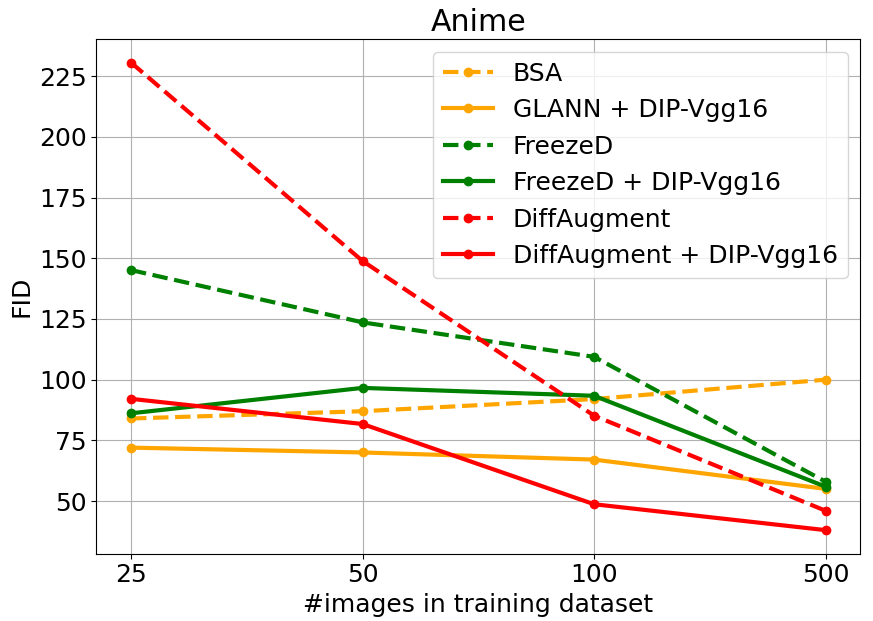}
  \caption{}\label{fig:vary_eg_anime}
\end{subfigure}
\caption{\footnotesize{\textbf{(a) and (b)}: Sample interpolations between two generated images for models trained in few-shot setting : Scratch \textit{(Row 1)}, Scratch + DISP-Vgg16 \textit{(Row 2)}, FreezeD \textit{(Row 3)}, FreezeD + DISP-Vgg16 \textit{(Row 4)}, DiffAugment \textit{(Row 5)}, DiffAugment + DISP-Vgg16 \textit{(Row 6)}. \textbf{(c)}: FID (lower is better) performance graph of few-shot image generation by varying the training samples from 25 to 500 images of Anime dataset for different approaches on SNGAN model. }}
\label{fig:few_shot_example_more}
\end{minipage} 
\end{figure*}

\begin{table}
\centering
\scalebox{0.8}{
\begin{tabular}{l r r r }
\hline
& \multicolumn{3}{c }{BigGAN (128 x 128)}  \\
\textbf{Method} & \textbf{Places2.5k } & \textbf{FFHQ2k} & \textbf{CUB6k}  \\
& FID $\downarrow$ & FID  $\downarrow$ & FID $\downarrow$
\\  \hline & &  &  \\
MineGAN & 75.50  &  75.91 & 69.64\\
\hline &    & &  \\
TransferGAN & 162.91  &  126.23 & 138.87 \\
+ DISP-Vgg16 & \textbf{57.35}  & \textbf{44.43} &  \textbf{23.37}\\
\hline &    & &  \\
FreezeD & 191.04  & 161.87 & 142.47 \\
+ DISP-Vgg16 & \textbf{50.58} & \textbf{43.90}   & \textbf{26.90} \\
\hline &    & &  \\
DiffAugment & 56.48 & 31.60 &  36.09 \\
+ DISP-Vgg16  & 30.76  & 23.19 &  15.81 \\
+ DISP-SimCLR  & \textbf{26.65} &  \textbf{21.06} & \textbf{12.36} \\
\hline 
\end{tabular}}
\captionof{table}{\footnotesize{FID of different techniques on limited data image generation. FID (lower is better) is computed using $10$k, $7$k, $6$k generated and real samples (disjoint from training set) for Places2.5k, FFHQ2k, CUB datasets respectively. All above approaches are initialized with BigGAN model pre-trained on ImageNet.}}

\label{table:limited-data}
\end{table}

\paragraph{Results} Using DISP shows consistent improvement in FID and Recall over all baseline methods as shown in Table \ref{table:fewshot}. Fig \ref{fig:few_shot_example_more} shows samples generated via interpolation between conditional embedding of models trained via DISP-Vgg on DiffAugment and vanilla DiffAugment. These results qualitatively show the improvement obtained using our DISP transfer learning approach. Comparatively, the baseline, vanilla DiffAugment, fails to generate realistic interpolation and for the most part, presents memorized training set images. DISP also performs better when training is done from scratch as compared to FreezeD and TransferGAN but is worse than DiffAugment + DISP. We present additional ablation studies in Supplementary.

\paragraph{Performance on varying number of training images} We vary the number of training examples in Anime dataset from $25$-$500$ for baseline few-shot algorithms and their respective augmentations with DISP-Vgg16. The FID metric comparison in Fig \ref{fig:vary_eg_anime} shows the benefit of our approach when used with existing training algorithms. The FID metric for all approaches improves (decreases) with the increase the number of training images with DISP out-performing corresponding baselines. Sample images generated by our approach are shown in Supplementary.

\paragraph{Memorization Test} To evaluate whether trained GANs are actually generating novel images instead of only memorizing the training set, we calculate FID between images randomly sampled from training set with repetition and the separate test set for Anime and FFHQ dataset. For Anime dataset, we get an FID of $81.23$ and for FFHQ, $100.07$. On comparing these numbers to Table \ref{table:fewshot} we observe that only on using DISP with existing algorithms, we achieve a better FID score suggesting that our approach is able to generate novel/diverse samples instead of just memorizing or over-fitting to training data. 

\paragraph{Analyzing the feature space of Vgg-16/SimCLR pre-trained network for Anime dataset} To examine the usefulness of Vgg features on Anime dataset, we evaluate it on the anime character classification task. We took a subset of 70k images from the Anime Face dataset that had labels assigned among the 50 character tags. Each character tag has around 1000-1500 images. We train a single linear classifier on Vgg-16 features of 50k samples and evaluate it on the rest 20k samples. We observe an accuracy of ~75\% and ~67\% on training and test sets respectively. When a single linear classifier is trained upon SimCLR features, the respective accuracies were ~81\% and ~63.5\%. This highlights that even for fine-grained and out of domain distributions like Anime, pre-trained Vgg-16 features are semantically rich enough to achieve a decent classification score.

\subsection{Limited Data Image Generation}\label{sec:limited}
\begin{figure*}[t]
\centering
\scalebox{0.8}{
\begin{subfigure}{0.33\textwidth}
  \includegraphics[width=\linewidth]{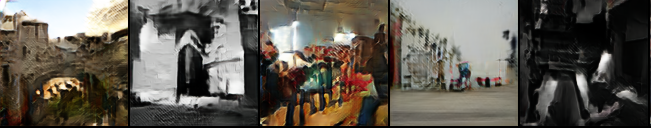}
  \includegraphics[width=\linewidth]{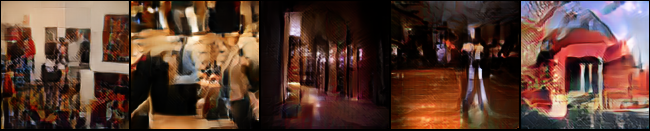}
  \includegraphics[width=\linewidth]{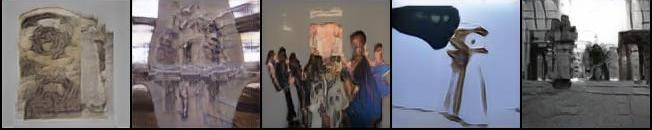}
  \includegraphics[width=\linewidth]{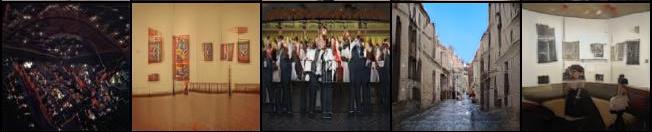}
  \caption{Places (2.5k)}
  \label{fig:2}
\end{subfigure} \hfil 
\begin{subfigure}{0.33\textwidth}
  \includegraphics[width=\linewidth]{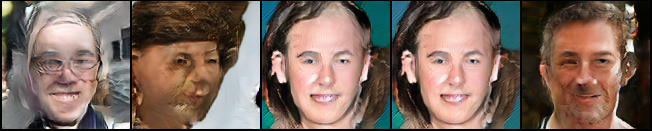}
  \includegraphics[width=\linewidth]{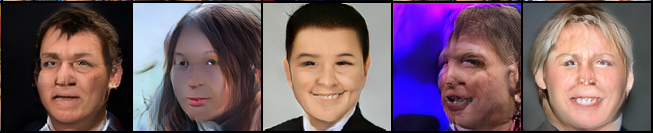}
  \includegraphics[width=\linewidth]{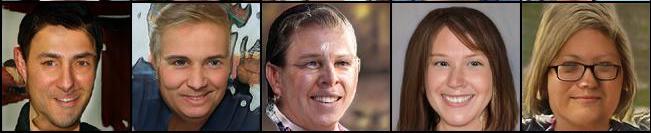}
  \includegraphics[width=\linewidth]{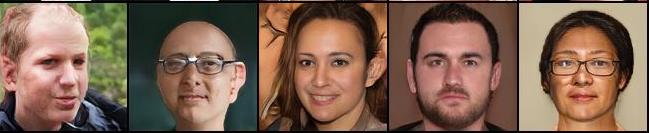}
  \caption{FFHQ (2K)}
  \label{fig:2}
\end{subfigure} \hfil 
\begin{subfigure}{0.33\textwidth}
  \includegraphics[width=\linewidth]{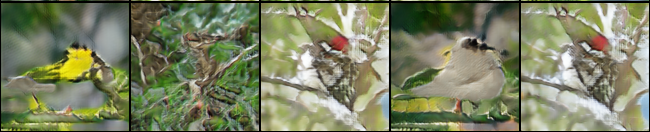}
  \includegraphics[width=\linewidth]{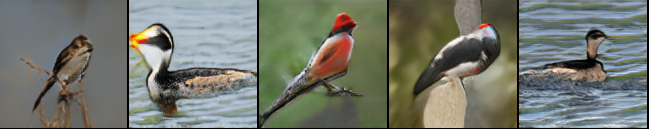}
  \includegraphics[width=\linewidth]{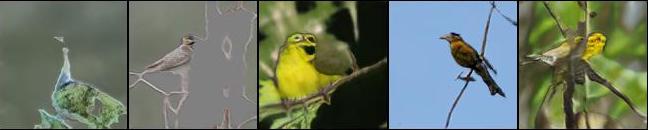}
  \includegraphics[width=\linewidth]{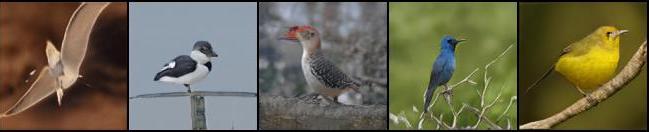}
  \caption{CUB (6K)}
  \label{fig:2}
\end{subfigure} 
}
\caption{\footnotesize{Samples of generated image in limited data training setting :  FreezeD \textit{(Row 1)}, FreezeD + DISP-Vgg16 \textit{(Row 2)}, DiffAugment \textit{(Row 3)} and DiffAugment + DISP-Vgg16 \textit{(Row 4)}.}}
\label{fig:limited-data}
\end{figure*}

In many practical scenarios, we have access to moderate number of images ($1$k-$5$k) instead of just a few examples, however the limited data may still not be enough to achieve stable GAN training. We show the benefit of our approach in this setting and compare our results with: MineGAN\cite{minegan_2020_CVPR}, TransferGAN, FreezeD, and DiffAugment. We perform experiments on three $128\times128$ resolution datasets: FFHQ2k, Places2.5k and CUB6k following \cite{minegan_2020_CVPR}. FFHQ2k contains 2K training samples from FFHQ \cite{ffhq_stylegan} dataset. Places2.5k is a subset of Places365 dataset \cite{places365} with $500$ examples each sampled from $5$ classes (alley, arch, art gallery, auditorium, ball-room). CUB6k is the complete training split of CUB-200 dataset \cite{cub}. We use the widely used BigGAN \cite{biggan} architecture, pre-trained on ImageNet for finetuning. Table \ref{table:limited-data} shows our results;  using DISP consistently improves FID on existing baselines by a significant margin. More implementation details are given in supplementary and sample generated images via our approach are shown in Figure \ref{fig:limited-data}. 
\vspace{-8pt}
\paragraph{Experiments on CIFAR-10 and CIFAR-100}
We also experiment with unconditional BigGAN and StyleGAN2 model on CIFAR-10 and CIFAR-100 while varying the amount of data as done in \cite{diffaug2020zhao}. We compare DISP with DiffAugment on all settings and the results are shown in Table \ref{table:fewshot_stylegan_cifar10-100}. In the limited data setting ($5\%$ and $10\%$) augmenting DiffAugment with DISP gives the best results in terms of FID for both BigGAN and StyleGAN2 architectures. When trained on complete training dataset DISP slightly outperforms DiffAugment on BigGAN architecture. For implementation details, please refer to supplementary.

\begin{table*}
    \centering
   \scalebox{0.8}{
    \begin{tabular}{l r r r r r r }
    \hline
    Method & \multicolumn{3}{c}{CIFAR-10} & \multicolumn{3}{c}{CIFAR-100}  \\
    & 100$\%$ data & 20$\%$ data & 10$\%$ data & 100$\%$ data & 20$\%$ data & 10$\%$ data \\
    \hline
    BigGAN  & 17.22 & 31.25 & 42.59 & 20.37 & 33.25 & 42.43 \\
    + DISP &  9.70 & 16.24 & 27.86 & 12.89 &  21.70 & 31.48 \\
    + DiffAugment & 10.39 & 15.12 & 18.56 & 13.33 & 19.78 & 23.80  \\
    + DiffAugment $\&$ DISP & \textbf{9.52}  & \textbf{14.24} & \textbf{18.50} & \textbf{12.70} & \textbf{16.91} & \textbf{20.47} \\
    \hline
    StyleGAN2$^{*}$ & 11.07 & 23.08 & 36.02 & 16.54 & 32.30 & 45.87\\
    + DiffAugment$^{*}$ & 9.89 & 12.15 & 14.5 & 15.22 & 16.65 & 20.75\\
    + DiffAugment $\&$ DISP & \textbf{9.50} & \textbf{10.92} & \textbf{12.03} & \textbf{14.45} & \textbf{15.52} & \textbf{17.33} \\
    \hline
    \end{tabular}}
    \caption{\footnotesize{Comparison of FID on Unconditional CIFAR-10 and CIFAR-100 image generation while varying the amount of training data. Here, all mentioned approaches are trained with random-horizontal flip augmentation of real images. BigGAN-DiffAugment includes consistency regularization \cite{crgan2019consistency} following the implementation provided by authors \cite{diffaug2020zhao}. We report the checkpoint with the best FID value for each model. $^{*}$ denotes directly reported from paper.}} \label{table:fewshot_stylegan_cifar10-100}
\end{table*}

\begin{table*}[t]
\centering

\scalebox{0.85}{
\begin{tabular}{|l|r r r|r r r|r r r|r r r|r r r|}
\hline
\textbf{Method}& \multicolumn{3}{c|}{\textbf{CIFAR-10}} & \multicolumn{3}{c|}{\textbf{CIFAR-100}} &   \multicolumn{3}{c|}{\textbf{FFHQ } }&  \multicolumn{3}{c|}{\textbf{LSUN-Bedroom}} &  \multicolumn{3}{c|}{\textbf{ImageNet32x32}} \\
& FID $\downarrow$ & P $\uparrow$ & R $\uparrow$ & FID $\downarrow$ & P $\uparrow$ & R $\uparrow$ & FID $\downarrow$ & P $\uparrow$ & R $\uparrow$ & FID $\downarrow$ & P $\uparrow$ & R $\uparrow$ & FID $\downarrow$ & P $\uparrow$ & R $\uparrow$ \\
\hline & & & & & & & & & & & & & & & \\
Baseline & 19.73 & 0.64 & \textbf{0.70} &  24.66 & 0.61 & \textbf{0.67} & 21.67 & 0.77 & 0.47 & 9.89 & 0.58 & 0.42 & 16.19 & 0.60 & \textbf{0.67} \\
SSGAN & 15.65 & 0.67 & 0.68  & 21.02 & 0.61 & 0.65 & - & - & - & 7.68 & 0.59 & 0.50 & 17.18 & 0.61  & 0.65 \\
Self-Cond GAN & 16.72 & 0.71 & 0.64  & 21.8 & 0.64 & 0.60 & - & - & - & - & - & -  & 15.56 &\textbf{0.66}  & 0.63 \\
\hline & & & & & & & & & & & & & & & \\
DISP-Vgg16 & \textbf{11.24} & \textbf{0.74} & 0.64 &  \textbf{15.71} & \textbf{0.70} & 0.62 & \textbf{15.83} & 0.76 & \textbf{0.55} & 4.99 & \textbf{0.66} & \textbf{0.54} & \textbf{12.11} & 0.64 & 0.62  \\
DISP-SimCLR  & 14.42 & 0.68 & 0.65 & 20.08 & 0.67 & 0.62 & 16.62 & \textbf{0.77} & 0.53 & \textbf{4.92} & 0.62 & 0.53 & 14.99 & 0.60  & 0.63  \\

\hline
\end{tabular}}
\caption{\footnotesize{Comparison of DISP with Baseline, SSGAN\cite{ssgan} and Self-Cond GAN\cite{liu2020selfconditioned} in large-scale image generation setting.
}}
\label{table:large-scale}
\end{table*}

\subsection{Large-Scale Image Generation}\label{sec:large_scale}
In order to show the usefulness of our method on large-scale image generation, we carry out experiments on CIFAR-10, CIFAR-100 \cite{krizhevsky2010cifar} and ImageNet-$32\times32$ datasets with $50$k, $50$k and $\sim$ $1.2$M training images respectively at $32\times32$ resolution. For a higher $128\times128$ resolution, we perform experiments on FFHQ and LSUN-bedroom \cite{lsun} datasets with $63$k and $3$M training samples. We use a ResNet-based architecture for both discriminator and generator similar to BigGAN \cite{biggan} for all our experiments. We also compare DISP with SSGAN \cite{ssgan} and Self-Conditional GAN \cite{liu2020selfconditioned}. Implementation and training hyperparameter details are provided in Supplementary. 

\begin{figure}
\centering
\includegraphics[width=\linewidth]{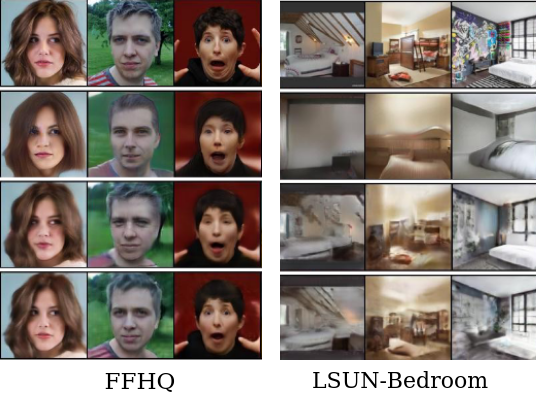}
\captionof{figure}{\footnotesize{Images generated through IvOM for randomly sampled test set images on FFHQ and LSUN-Bedroom. \textit{(Top to Bottom:)} Original images, Baseline, Baseline + DISP-Vgg16, Baseline + DISP-SimCLR.}}
\label{fig:inversion}
\end{figure}

\vspace{-8pt}
\paragraph{Results} Table \ref{table:large-scale} reports the FID, precision and recall score on the generated samples and the test set for baselines and our approach (DISP). For fitting GMM, the number of components are fixed to $1K$ for all datasets. DISP achieves better FID, precision and recall scores compared to leading baselines. Sample qualitative results and generation with latent interpolation are shown in the supplementary. We also evaluate the quality of inverted images for $128\times128$ resolution on FFHQ and LSUN datasets using Inference via Optimization Measure (IvOM) \cite{ivom2017metz} to emphasize the high instance-level data coverage in the prior space of GANs trained through our approach (details on IvOM calculation are provided in supplementary). Table \ref{table:inversion} shows the IvOM and FID metric between inverted and real query images. Figure \ref{fig:inversion} shows sample inverted images. We observe both from qualitative and quantitative perspective, models trained via DISP inverts a given query image better than the corresponding baselines. We also perform an ablation experiment to analyze the effect of different priors in DISP for CIFAR-100 dataset. As shown in Table \ref{table:otherprior}, the FID metric remains relatively similar for different priors when compared to the baseline. 
\vspace{-8pt}
\paragraph{Memorization test} For analyzing memorization in GANs, we evaluate it on the recently proposed test to detect data copying \cite{datacopy}. The test calculates whether generated samples are closer to the training set as compared to a separate test set in the inception feature space using three sample Mann-Whitney U test \cite{mann1947test}. The test statistic $C_T << 0$ represents overfitting and data-copying, whereas $C_T >> 0$ represents underfitting. We average the test statistic over $5$ trials and report the results in Table \ref{table:datacopy}. We can see that using data instance priors during GAN training does not lead to data-copying according to the test statistic except in case of FFHQ dataset where both DISP and baseline $C_T$ values are also negative.  \par

\textbf{Performance gain due to knowledge distillation vs Memorization of real image features} We conduct an additional experiment where we use the features of a Resnet50 network trained on $75$\% label-corrupted CIFAR-100 as Data Instance Priors to train CIFAR-100 BigGAN architecture.  This results in a significantly higher FID ($22.82$) in comparison to using prior feature from Resnet50-SimCLR trained on clean CIFAR100 dataset,(FID $14.62$, Table \ref{table:otherprior}).
This highlights that performance depends on the quality of pre-trained network features and not only because features of real images are leveraged as prior during generation. If this was not the case then using features of Resnet50 trained on $75$\% label-corrupted CIFAR-100 would have resulted in similar performance.

\begin{table}[t]
    \centering
      \scalebox{0.8}{
  \begin{tabular}{ l|rrrr}
    \hline
    \textbf{Method}&  \textbf{50k} & \textbf{100k} & \textbf{200k} &	\textbf{500k} \\
    \hline
    GMM &	4.99 &	4.92 &	4.81 &	4.43 \\
Time (s) & 	383.96 &	1063.99 &	1993.93 &	4397.56 \\ \hline
    \end{tabular}}
    \caption{\footnotesize{Relationship among the number of random samples used in the GMM, FID value obtained and the time taken for learning the GMM.}} \label{table:clustering_time}
\end{table}

\begin{table}
\centering
\scalebox{0.7}{
\begin{tabular}{|l|r| r| r| r | r| }
    \hline
    \textbf{Methods}  & \textbf{CIFAR-10} & \textbf{CIFAR-100} & \textbf{FFHQ} & \textbf{LSUN} & \begin{tabular}{@{}c@{}}\textbf{ImageNet} \\ \textbf{32x32}\end{tabular} \\
     &  $C_T$  &  $C_T$ & $C_T$ &  $C_T$ &  $C_T$
    \\  \hline & &  &  & & \\
    Baseline &3.02 & 4.26 & -0.15 & 2.59 & 10.5 \\ 
    DISP-Vgg16 & 1.58 & 3.05 & -0.81 & 1.06 & 8.53 \\ 
    DISP-SimCLR & 2.86 & 3.48 &  -1.49 & 0.13 & 9.91 \\
    \hline
    \end{tabular}}
    \captionof{table}{\footnotesize{Test for evaluating data-copy and memorization in GANs \cite{datacopy} for different approaches and datasets. Test statistic $C_T << 0$ denotes overfitting and data-copying, and $C_T >> 0$ represents under-fitting.}} \label{table:datacopy}
\end{table}

\paragraph{Relationship between the number of random samples used for fitting GMM and its corresponding FID} Fitting a GMM model is inhibitive for large-scale datasets e.g. ImageNet and LSUN-Bedroom where the training data is in millions. We observed this during our experiments and therefore use a subset of randomly sampled 200K instances for these datasets for fitting GMM in Table \ref{table:large-scale}. Table \ref{table:clustering_time} further shows the relationship between the number of random samples used for fitting GMM and the corresponding FID (average of 3 runs with a standard deviation of less than 1\%) on the LSUN-Bedroom dataset for DISP-Vgg16 trained model on LSUN-Bedroom. As can be seen, even a small subset of training data can still be used to achieve better performance than baselines in relatively less time. This experiment was performed on a system with 32 CPU cores, 64 GB RAM, and processor Intel(R) Xeon(R) CPU @ 2.20GHz.

\begin{table}[]
    \centering
 \scalebox{0.8}{
    \begin{tabular}{|l|c|c|c|c|c|c|}
    \hline
    \textbf{Method}&    \multicolumn{2}{c|}{\textbf{FFHQ } }&  \multicolumn{2}{c|}{\textbf{LSUN-Bedroom}} \\
    & IvOM $\downarrow$ & FID $\downarrow$ & IvOM $\downarrow$ & FID $\downarrow$ \\
    \hline & & & & \\
    Baseline & 0.0386 & 85.06 & 0.0517 & \textbf{115.02}  \\
    + DISP-Vgg16 & 0.0142 & 73.85 & 0.0191  & 129.4 \\
    + DISP-SimCLR & \textbf{0.0125} & \textbf{71.44} & \textbf{0.0161} & 116.11 \\
    \hline
    \end{tabular}}
    \captionof{table}{\footnotesize{IvOM and FID measure on 500 random test images of FFHQ and LSUN-Bedroom datasets.}}
    \label{table:inversion}
\end{table}

\begin{table}
    \centering
  \scalebox{0.8}{
  \begin{tabular}{ |l|c|}
    \hline
    \textbf{Method}& \textbf{CIFAR-100}\\
    \hline
    Baseline &  24.66  \\
    + DISP-SimCLR \small(ImageNet) &  16.26		\\
    + DISP-SimCLR \small(CIFAR-100) & 14.62 \\
    + DISP-ResNet50 \small(Places-365) & 14.68 \\
    + DISP-Resnet50 (ImageNet) & 14.62 \\
    \hline
    \end{tabular}}
    \captionof{table}{\footnotesize{Comparison of FID when using prior from different pre-trained models on CIFAR-100.}} \label{table:otherprior}
\end{table}

\section{Conclusion}
In this work, we present a novel instance level prior based transfer learning approach to improve the quality and diversity of images generated using GANs when a few training data samples are available. By leveraging features as priors from rich source domain in limited unsupervised image synthesis, we show the utility of our simple yet effective approach on various standard vision datasets and GAN architectures. We demonstrate the efficacy of our approach in image generation with limited data, where it achieves the new state-of-the performance, as well as on large-scale settings. As future work, it would be interesting to explore the application of prior information in image editing tasks.

{\small
\bibliographystyle{ieee_fullname}
\bibliography{egbib}
}

\twocolumn[\appendixhead]

\section{Few-Shot Image Generation}

\paragraph{Few-shot image generation (StyleGAN2)} We follow the experimental setting of \cite{diffaug2020zhao} and show performance on 100-shot Obama, Panda and Grumpy Cat datasets (having 256 $\times$ 256 resolution) using FFHQ \cite{ffhq_stylegan} pre-trained StyleGAN2 model. 
Table \ref{table:stylegan} shows DISP training leads to consistent improvement in FID scores over several baseline techniques except on Grumpy Cat dataset. We hypothesize that this is because the prior features of this dataset has low diversity and hence the priors used are not informative enough to lead to improved performance with DISP. 

\begin{table}[!h]
\centering
    \scalebox{0.9}{
    \begin{tabular}{l r r r}
    \hline
    & \multicolumn{3}{c}{Style-GAN 2 (256 x 256)}  \\
    \multirow{2}{*}{\textbf{Method}} & \textbf{Panda } & \textbf{Grumpy Cat} & \textbf{Obama} \\
    & FID $\downarrow$ & FID $\downarrow$ & FID $\downarrow$
    \\  \hline & &  &  \\
   
    FreezeD & 16.69 & \textbf{29.67} & 62.26 \\
    + DISP-Vgg16 & \textbf{14.66} & 29.93  & \textbf{54.87} \\
    \hline  & &  &  \\
    DiffAugment & 12.06 & \textbf{27.08}  & 46.87  \\
    + DISP-Vgg16 & \textbf{11.14} & 28.45 & \textbf{43.79} \\
    \hline  & &  &   \\
    BSA* & 21.38 & 34.20 & 50.72 \\
    GLANN + DISP-Vgg16 & \textbf{11.51}  &  \textbf{29.85} & \textbf{38.57} \\
    \hline
    \end{tabular}}

\caption{\footnotesize{ $100$-shot image generation results using StyleGAN2 \cite{stylegan2} model pre-trained on FFHQ dataset for Panda, Grumpy-cat and Obama datasets. FID is computed between $5$k generated and the complete training dataset. * denotes directly reported from the paper \cite{diffaug2020zhao}.}}
\label{table:stylegan}
\end{table}

\paragraph{Impact of loss function} To analyze the role of GAN loss function, we show the performance of DISP with different variants. Specifically, we choose these three loss functions: hinge loss (originally in our experiments), non-saturating loss \cite{gan_goodfellow} and the wasserstein loss \cite{arjovsky2017wasserstein}. Table \ref{table:fewshot_lossfn} shows the corresponding results when DISP is used with FreezeD and DiffAugment. We observe that in case of FreezeD+DISP wasserstein loss significantly outperforms non-saturating loss and hinge loss. In case of DiffAugment hinge loss performs best followed by non-saturating loss and wasserstein loss.   

\begin{figure}[!h]
\centering 
  \includegraphics[width=\linewidth]{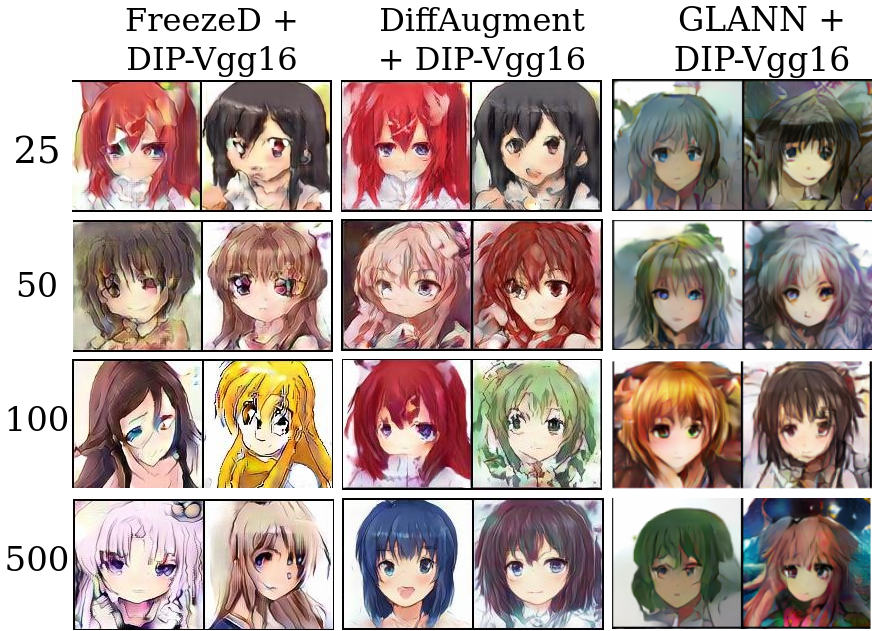}
\caption{\footnotesize{Samples of few-shot image generation on varying the number of training data from $25$ to $500$ images of Anime dataset using DISP along with different approaches with SNGAN backbone.}}
\label{fig:vary_eg_anime_examples}
\end{figure}

\paragraph{Samples by varying number of training images} Figure \ref{fig:vary_eg_anime_examples} shows samples generated by our approach when we vary the number of training examples in Anime dataset from 25-500. For quantitative results please refer Figure 3c in main submission.
\begin{table}[t]
\centering
\scalebox{0.7}{
    \begin{tabular}{l r r r r r r r}
    \hline
    & & \multicolumn{6}{c}{SNGAN (128 x 128)}   \\
    \multirow{2}{*}{\textbf{Method}} & \textbf{\begin{tabular}{@{}c@{}}Pre- \\ training\end{tabular}} &  \multicolumn{3}{c}{\textbf{Anime}} &  \multicolumn{3}{c}{\textbf{Faces}} \\ 
    &  & H  & NS  & W  & H  & NS  & W  \\  
    \hline &    &  &  & &  &  &  \\
  
    FreezeD & $\checkmark$ &109.40  & 102.43 & 148.99 & 107.83 & 105.34 &209.23 \\
    + DISP-Vgg16 & & 93.36 & 82.49  & 74.91 & 77.09  & 77.38 & 71.05 \\
   
    \hline & &  &    & & &  &  \\
    DiffAugment & $\times$ & 85.16  & 106.96  & 252.11 &  109.25 & 107.18  & 325.85   \\
    + DISP-Vgg16 & & 48.67   & 48.61 & 56.43 &  62.44  & 68.66 & 81.03  \\
    
    \hline &    &   & & &  &  &  \\
    
    \end{tabular}}
    \caption{\footnotesize{Comparison between different loss functions in few-shot image generation using 100 training images (FID: lower is better). H is hinge loss, NS is non saturating loss and W is wasserstein loss.}} \label{table:fewshot_lossfn}
\end{table}

\SetAlgoLined
\begin{algorithm}[!t]
\footnotesize{
\textbf{Input}:$G$, $D$ network with parameters $\theta_G$ and $\theta_D$, pre-trained model $C$ for extracting prior condition, samples from real data distribution $q(x)$ and latent distribution $p(z)$, batch size $b$, number of training iterations, discriminator update steps $d_{step}$ for each generator update, Adam optimizer hyperparameters $\alpha, \beta_1 , \beta_2$. 
  
\For{\text{number of training iterations}}
    {
        \For{$t:1 ... d_{\textit{step}}$}
            {   Sample batch $x \sim q(x) $, $ z  \sim p(z) $ \\
              $x_{\textit{fake}} =G(z|C(x)) $\\ 
              
              $D(x , C(x)) = D_f(x) \boldsymbol{\cdot} D_{emb}(C(x)) + D_l \circ D_f(x)$\\
              $D(x_{\textit{fake}},C(x)) =  D_f(x_{\textit{fake}}) \boldsymbol{\cdot} D_{emb}(C(x)) + D_l \circ D_f(x_{\textit{fake}})$\\
              $L_{D} = \max(0, 1-D(x, C(x))) + \max(0, 1+D(x_{\textit{fake}}, C(x)))$ \\
              Update $\theta_{D} \leftarrow
                Adam( L_{D} , \alpha, \beta_1 , \beta_2) $
            }
        Sample $ z  \sim p(z) $ \\
        Generate images $x_{\textit{fake}} = G(z|C(x)$ \\ 
        $D(x_{\textit{fake}},C(x)) = D_f(x_{\textit{fake}}) \boldsymbol{\cdot} D_{emb}(C(x)) +  D_l \circ D_f(x_{\textit{fake}}) $\\
        $L_G =  - D(x_{\textit{fake}}, C(x))$  \\
        Update $\theta_{G} \leftarrow
                Adam( L_{G} , \alpha, \beta_1 , \beta_2) $
    }
\textbf{return} $\theta_G, \theta_D $.}
\caption{\footnotesize{Data InStance Prior (DISP) training algorithm}}
\label{train_sal}\label{algo:mainalgo_dip}
\end{algorithm}

\paragraph{Implementation Details}\label{app:fewshot_implementation_details} 
We summarize the training procedure of DISP in Algorithm \ref{algo:mainalgo_dip}.

In SNGAN architecture, while training with DISP, $G_{emb}$ and $D_{emb}$ are matrices which linearly transform the pre-trained features into generator conditional space of dimension $128$ and discriminator feature space of dimension $1024$. For baseline training, we use an embedding for each of the $100$ training images to ensure minimal difference between baseline and our approach without increasing number of parameters. We also experimented with self-modulated \cite{selfmod} and unconditional training which resulted in either training collapse or worse results in all approaches. In DiffAugment, we use three augmentations: translation, cutout, and color with consistency regularization hyperparameter as $10$ and training is done from scratch following the implementation in their paper \cite{diffaug2020zhao}. In FreezeD, we freeze the first five blocks of the discriminator and finetune the rest. We use spectral normalization for both generator and discriminator during training with batch size of $25$, number of discriminator steps as $4$, $G$ and $D$ learning rate as $2e-4$, $\mathbf{z}$ dimension as $120$ and maximum number of training steps as $30K$. During evaluation, moving average weights \cite{inception2016} of the generator is used in all experiments unless stated otherwise. For FID calculation, we select the snapshot with best FID similar to \cite{ssgan,diffaug2020zhao}. For calculating precision and recall based on the k-nearest neighbor graph of inception features, as in \cite{kynkaanniemi2019improved_pr}, we use $k$ as $10$ for Precision and $40$ for Recall.

For StyleGAN2, $G_{emb}$ is a 2-layer MLP with ReLU non-linearity which maps $C(\mathbf{x})$ to a $512$-dimensional generator conditional space. It is then concatenated with random noise $\mathbf{z}$ of dimension $512$ which is used as input in the mapping network. $D_{emb}$ is a linear transformation matrix and discriminator loss is projection loss combined with real/fake loss. Training is done with batch-size of $16$ for DiffAugment\footnote{https://github.com/mit-han-lab/data-efficient-gans} and $8$ for FreezeD\footnote{https://github.com/sangwoomo/FreezeD} till $20k$ steps.   

In case of BSA, we show that DISP can be used to improve the results on similar non-adversarial generative models. Specifically, we perform experiments with GLANN \footnote{https://github.com/yedidh/glann} which is a two step training procedure, as follows: (1) Optimize for image embeddings $\{\mathbf{e}_i\}$ of all training images $\{\mathbf{x}_i\}$ jointly with a generator network $G$ using perceptual loss; and (2) Learn a sampling function $T:\mathbf{z} \rightarrow \mathbf{e} $ through IMLE for generating random images during inference. For using data instance prior in the training procedure of GLANN, instead of directly optimizing for $\{\mathbf{e}_i\}$, we optimize for the following modified objective: 
\begin{equation}
    \begin{aligned}
     \underset{G , G_{emb} }{\arg \min} & \sum_i L_{perceptual} (G \circ G_{emb} \circ C (\mathbf{x}_i) , \mathbf{x}_i)\\
     \textnormal{where} \;\;& \{e_i\} = \{ G_{emb} \circ C(\mathbf{x}_i)\}
    \end{aligned}
\end{equation}
We finetune the pre-trained generator on batch-size of $50$ with a learning rate of $0.01$ for 4000 epochs. During second step of IMLE optimization, we use a 3-layer MLP with $\mathbf{z}$ dimension as $64$ and train for $500$ epochs with a learning rate of $0.05$.

\paragraph{Comparison with Logo-GAN} Logo-GAN \cite{sage2018logo} has shown advantage of using features from pre-trained ImageNet network in unconditional training by assigning class label to each instance based on clustering in the feature space. We compare our approach with this method in the few-shot data setting. For implementing logo-GAN, we perform class-conditional training \cite{sngan} using labels obtained by K-means clustering on Vgg16 features of $100$-shot Anime dataset. The results reported in Table \ref{table:class_conditional_training_anime} show the benefit of directly using features as data instance prior instead of only assigning labels based on feature clustering. 

\begin{table}[!h]
    \centering
    \scalebox{0.9}{
\begin{tabular}{l r }
\hline
\textbf{Method} & \textbf{Anime (SNGAN)} \\
& FID $\downarrow$ \\  
\hline &   \\
FreezeD + DISP & \textbf{93.36}   \\
FreezeD + Logo-GAN (K=5) & 226.60 \\
FreezeD + Logo-GAN (K=10) & 183.38  \\
\hline & \\
DiffAugment + DISP  & \textbf{48.67}  \\
DiffAugment + Logo-GAN (K=5) & 130.54  \\
DiffAugment + Logo-GAN (K=10) & 190.59  \\
\hline 
\end{tabular}}
\caption{\footnotesize{$100$-shot image generation comparison of DISP with Logo-GAN \cite{sage2018logo} on Anime dataset where priors are derived from Vgg16 network trained on ImageNet. FID is computed between $10$k generated and real samples (disjoint from training set). }}
\label{table:class_conditional_training_anime}
\end{table}

\section{Limited data Image Generation}\label{app:limited}

\paragraph{Experiments on CIFAR-10 and CIFAR-100}
For results shown in Table 3 of main submission, BigGAN model used for training CIFAR-10 and CIFAR-100 is same as the one used for large scale experiments in Section 5.3 of main submission. In DiffAugment with BigGAN architecture, we use all three augmentations: translation, cutout, and color along with consistency regularization hyperparameter as $10$. In DiffAugment + DISP consistency regularization hyperparameter is changed to $1$. For experiments on StyleGAN2 architecture we use the code-base of DiffAugment \footnote{https://github.com/mit-han-lab/data-efficient-gans/tree/master/DiffAugment-stylegan2}.

\paragraph{Implementation details of experiment on 128 Resolution datasets with BigGAN architecture in Section 5.2 of main submission}
We use our approach in conjunction with existing methodologies in a similar way as the few-shot setting with $G_{emb}$ and $D_{emb}$ as linear transformation matrices which transform the data priors into the generator's conditional input space of dimension 128 and discriminator feature space of dimension $1536$. During baseline training, we use self-modulation \cite{selfmod} in the batch-norm layers similar to \cite{ssgan,unet}. In DiffAugment, we use three augmentations: translation, cutout, and color with consistency regularization hyperparameter as $10$. During FreezeD training, we freeze the first $4$ layers of discriminator. For TransferGAN, FreezeD, MineGAN and its augmentation with DISP, we use the following hyperparameter setting: batch size $256$, $G$ and $D$ lr $2e-4$ and $\mathbf{z}$ dimension $120$. For DiffAugment, batch size is $32$, D-steps is $4$ and rest of the hyperparameters are same. Training is done till $30$k steps for DiffAugment, FreezeD, and $5$k steps for the rest. The moving average weights of the generator are used for evaluation. We use pre-trained network from \footnote{https://github.com/ajbrock/BigGAN-PyTorch} \cite{biggan} for finetuning.

\section{Large-Scale Image Generation}\label{app:large_scale}

\begin{figure*}[t]
\begin{subfigure}[t]{0.49\textwidth}
\includegraphics[width=\linewidth]{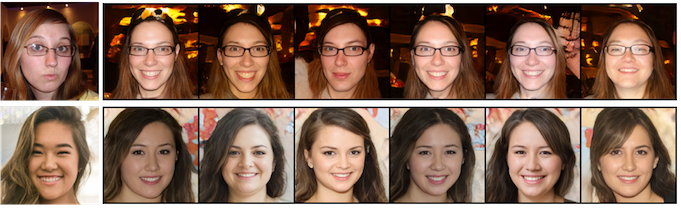}
\includegraphics[width=\linewidth]{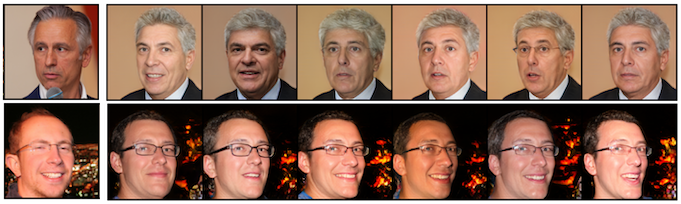}
\caption{Semantic Variations - First column corresponds to image whose Vgg16 features are given as prior to DISP module. Rest columns correspond to images generated using random noise. As can be seen the generated images are consistent with the prior image in terms of high-level semantics. }
\end{subfigure}%
\hfil
\begin{subfigure}[t]{0.49\textwidth}
\includegraphics[width=\linewidth]{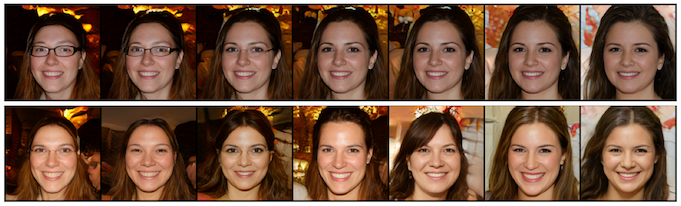}
\includegraphics[width=\linewidth]{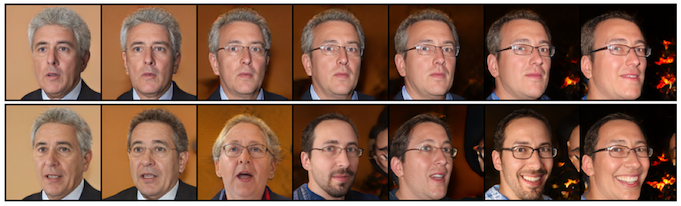}
\caption{Interpolation - First and last column corresponds to images used for interpolation. Rest columns correspond to images generated when Vgg16 features of images in first and last column are interpolated and given as prior to DISP module.}
\end{subfigure}
\par\medskip 

  \caption{\footnotesize{Semantic variations and interpolation of generated samples using pre-trained Vgg16 conditional DISP module on FFHQ dataset. \textit{(Left (top and bottom):)} Random samples generated with prior as feature of the first column of images in each row; \textit{(Right (top and bottom):)} all 4 rows show interpolation of images between the generated image in the first and last column.}}
\label{fig:semantic-variations}
\end{figure*}

\begin{table}[t]
    \centering
      \scalebox{0.9}{
  \begin{tabular}{ l|ccc}
    \hline
    \textbf{Pearson Correlation}&  \textbf{Anime} & \textbf{FFHQ} & \textbf{CIFAR-10} \\
    \hline
    $D_f$ cosine vs VGG Perceptual 	& 0.65 	& 0.81 &	0.80 \\
    $D_f$ cosine vs Image $L_2$ & -0.46 & -0.61 & -0.54 \\
    \hline
    \end{tabular}}
    \caption{\footnotesize{Pearson Correlation between cosine similarity in Discriminator feature space ($D_f$) vs Vgg-16 perceptual similarity/ $L_2$ distance in Image space on Anime, FFHQ and CIFAR-10 dataset}} \label{table:pearson_corr}
\end{table}
\begin{table}[!h]
    \centering
      \scalebox{0.9}{
  \begin{tabular}{ l|cc}
    \hline
    \textbf{Cosine Similarity}&  \textbf{$x$ and $G(z|C(x))$} & \textbf{Random pair} \\
    \hline
    VGG perceptual space & 0.512 $\pm$ 0.067 & 0.382
 $\pm$ 0.050 \\
    Discriminator’s feature space &  0.59 $\pm$  0.096 & 0.50 $\pm$  0.070 \\
    \hline
    \end{tabular}}
    \caption{\footnotesize{Similarity between $x$ and $G(z|C(x))$ vs Similarity between a random pair of images from FFHQ dataset.}} \label{table:cosine_sim}
\end{table}

\paragraph{Image inversion} \label{inversion_appendix}
To invert a query image, $\mathbf{x}_q$ using our trained model, we optimize the prior (after passing it to $G_{emb}$) that is used to condition each resolution block, independently. Mathematically, we optimize the following objective:
\[ \mathbf{z}^*, C_1^*, .. C_k^* = arg \min_{\mathbf{z},C_1, .. C_2} \Vert G(\mathbf{z}\vert C_1, .. C_k) - \mathbf{x}_q \Vert^2_{2} \text{, } \]
\[ \mathbf{x}^{inv}_q = G(\mathbf{z}^*|C_1^*, .. C_k^*) \]
Here, $C_i$ (after passing it through $G_{emb}$) is the prior that is used to condition the $i^{th} \in \{1...k\}$ resolution block.
To get a faster and better convergence, we initialize all $C_i$ as $G_{emb}(C(\mathbf{x}_q))$. The optimization is achieved via back-propagation using Adam optimizer with learning rate of 0.1. Figure 5 (main submission) shows sample inverted images on FFHQ and LSUN-Bedroom datasets. From the figure, we can see that models trained via DISP invert a given query image better than the corresponding baselines.

\paragraph{Equivalence of closeness in latent and image space} In our algorithm, we use projection loss in discriminator latent space $D_f$ to enforce that a generated image $G(z|C(x)$ is semantically similar/close to a given image $x$. And to verify if discriminator latent space is indeed good space to measure similarities, we measure the correlation between cosine similarity in Discriminator feature $D_f$ and Vgg-16 feature (perceptual similarity) space. Vgg-perceptual similarity is an accepted measure of image similarity and has been used in generative models like IMLE, GLANN, BSA as a proxy for constraints in image space. Additionally, we also report the correlation between cosine similarity in Discriminator feature space and $L_2$ closeness measure in the image space. Table \ref{table:pearson_corr} reports our findings where we observe a high positive correlation between cosine similarity in $D_f$ and VGG perceptual similarity; and a moderate negative correlation between cosine similarity $D_f$ in and $L_2$ distance in Image space.

To quantitatively verify that $G(z|C(x))$ is close to $x$ in the trained model, we also show in Table \ref{table:cosine_sim}, the perceptual similarity between the two as compared to a random pair of images from FFHQ dataset. We can observe that $x$ and $G(z|C(x)$ are more similar than any random pair of images. 

\begin{figure*}[t]
\begin{subfigure}[t]{0.48\textwidth}
\includegraphics[width=\linewidth, height=210pt]{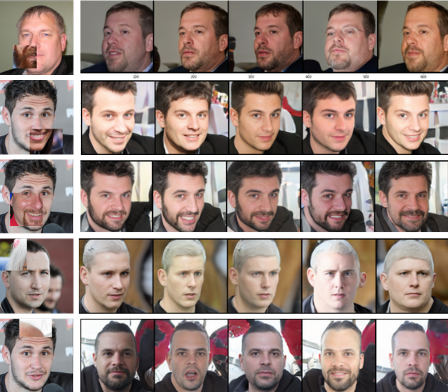}
\caption{Custom Editing - First column shows human-edited version where certain portion of image is substituted with another to achieve desired semantics. Rest columns correspond to images generated when Vgg16 features of human-edited version is provided as prior to DISP module.}
\end{subfigure}%
\hfil
\begin{subfigure}[t]{0.48\textwidth}
\includegraphics[width=\linewidth,height=210pt]{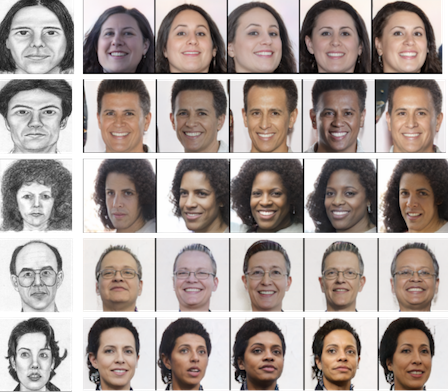}
\caption{Sketch-to-Image - First column shows sketch describing desired high-level semantics. Rest columns correspond to images generated when Vgg16 features of the sketch version is provided as prior in DISP module.}
\end{subfigure}
\par\bigskip 
\begin{subfigure}[t]{0.48\textwidth}
\includegraphics[width=\linewidth,height=210pt]{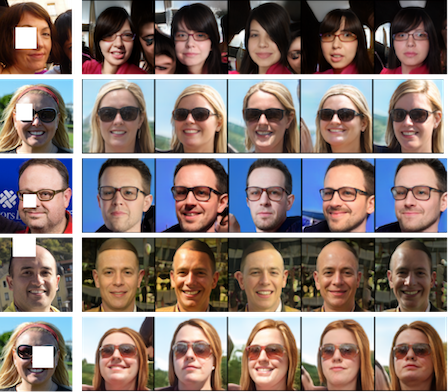}
\caption{Inpainting - First column shows a cutout in an Image. Rest columns correspond to images generated when Vgg16 features of the cutout version is provided as prior in DISP module.}
\end{subfigure}%
\hfil
\begin{subfigure}[t]{0.48\textwidth}
\includegraphics[width=\linewidth,height=210pt]{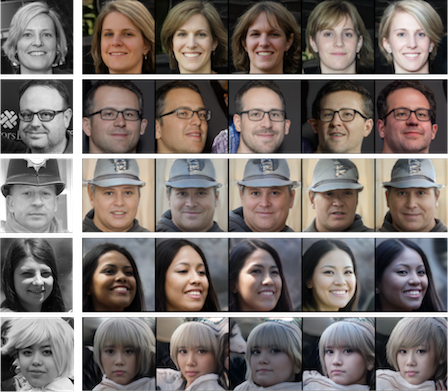}
\caption{Colourization - First column shows gray-scale image describing desired high-level semantics. Rest columns correspond to images generated when Vgg16 features of the gray-scale version is provided as prior in DISP module.}
\end{subfigure}
\caption{Examples of semantic diffusion used in image manipulation on FFHQ dataset using our DISP-Vgg16 approach. \textit{Top-Left:} Custom Editing; \textit{Top-Right:} Sketch-to-Image; \textit{Bottom-Left:} Inpainting;  \textit{Bottom-Right:} Colorization}
\label{app:cutmix}
\end{figure*}

\paragraph{Implementation Details} We use a single linear layer to transform the pre-trained image features to the generator's conditional input space of $128$ dimensions, and discriminator feature space of $1024$ dimensions respectively. A hierarchical latent structure similar to \cite{biggan} is used during DISP training. During evaluation with K-means and GMM on ImageNet and LSUN-Bedroom we first randomly sample 200K training images and then fit the distribution since clustering on complete training set which is in the order of millions is infeasible. In the training of the unconditional baseline, we use self-modulation \cite{selfmod}. In SSGAN, for rotation loss we use the default parameter of $0.2$ for generator and $1.0$ for discriminator as mentioned in \cite{ssgan}. For training Self-Conditional GAN \cite{liu2020selfconditioned}, we set the number of clusters to $100$ for all datasets. For CIFAR-10 and CIFAR-100, we re-cluster at every $25$k iterations with $25$k samples, and for ImageNet, at every $75$k iterations with $50$k samples following default implementation as in \cite{liu2020selfconditioned}. Following standard practice \cite{crgan2019consistency}, we calculate FID, Precision and Recall between test split and an equal number of generated images for-10, CIFAR-100, and ImageNet $32\times32$,  i.e., $10$k, $10$k, and $50$k, respectively. For FFHQ and LSUN-bedroom datasets, we use $7$k and $30$k generated and real (disjoint from training) samples, respectively. For all datasets and methods, training is done with batch size of $64$, G and D learning rate is set to $0.0002$, $\mathbf{z}$ dimension equals $120$ and spectral normalization is used in both generator and discriminator networks. Training is done till $100$k steps for all datasets except ImageNet which is trained for $200$k steps and moving average weights of generator are used during evaluation.

\paragraph{Semantic diffusion for image manipulation} \label{app:editing_more_examples}
We observed that high-level semantics (e.g. hair, gender, glasses, etc in case of faces) of a generated image, $G(\mathbf{z}\vert C(\mathbf{x}))$, relied on the conditional prior, $C(\mathbf{x})$. Complementarily, variations in the latent code $\mathbf{z} \sim \mathbf{N}(0, I)$ induced fine-grained changes such as skin texture, face shape, etc. This suggests that we can exploit conditional prior, $C(\mathbf{x})$, to get some control over the high-level semantics of generated image. We show that by altering an image $\mathbf{x}$ (through CutMix, CutOut, etc) and using $C(\mathbf{x})$ of the altered image as our new input prior helps in generating samples with the desired attributes, as shown in Fig \ref{app:cutmix}. In a similar manner, DISP also allows generation of images with certain cues (like sketch to image generation, as shown in Fig \ref{app:cutmix}). The generation of samples in this case is simply done by using $C(\mathbf{x})$ as prior in $G$.

\begin{figure*}[t]
    \centering 
\begin{minipage}{\linewidth}
 \centering
\begin{subfigure}{0.3\textwidth}
  \includegraphics[width=\linewidth]{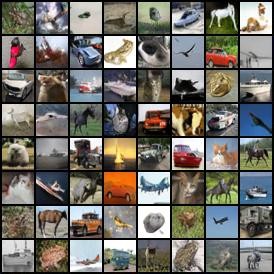}
  \caption{CIFAR-10}
  \label{fig:1}
\end{subfigure}\hfil 
\begin{subfigure}{0.3\textwidth}
  \includegraphics[width=\linewidth]{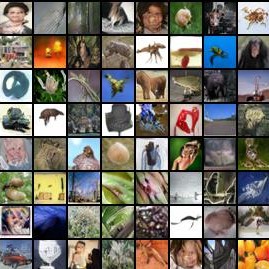}
  \caption{CIFAR-100}
  \label{fig:2}
\end{subfigure}\hfil 
\begin{subfigure}{0.3\textwidth}
  \includegraphics[width=\linewidth]{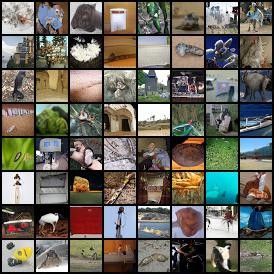}
  \caption{ImageNet-32x32}
  \label{fig:3}
\end{subfigure}
\begin{subfigure}{0.4\textwidth}
  \includegraphics[width=\linewidth]{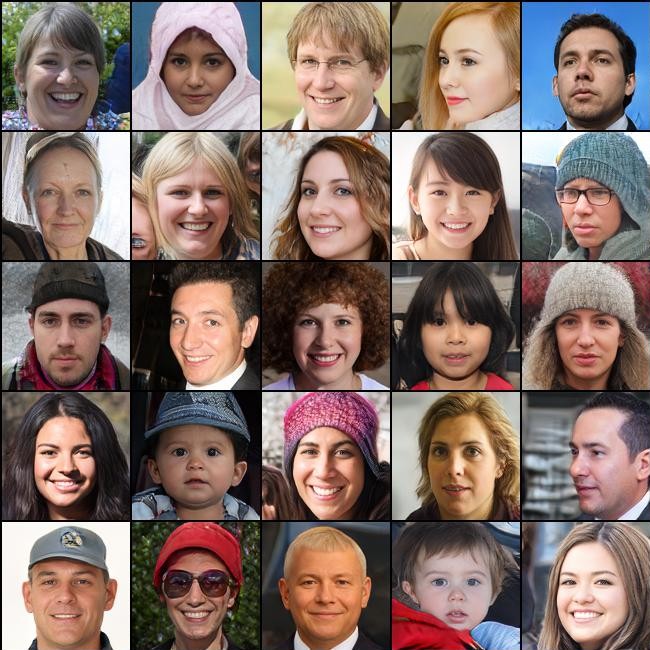}
  \caption{FFHQ}
  \label{fig:5}
\end{subfigure}\hfil 
\begin{subfigure}{0.4\textwidth}
  \includegraphics[width=\linewidth]{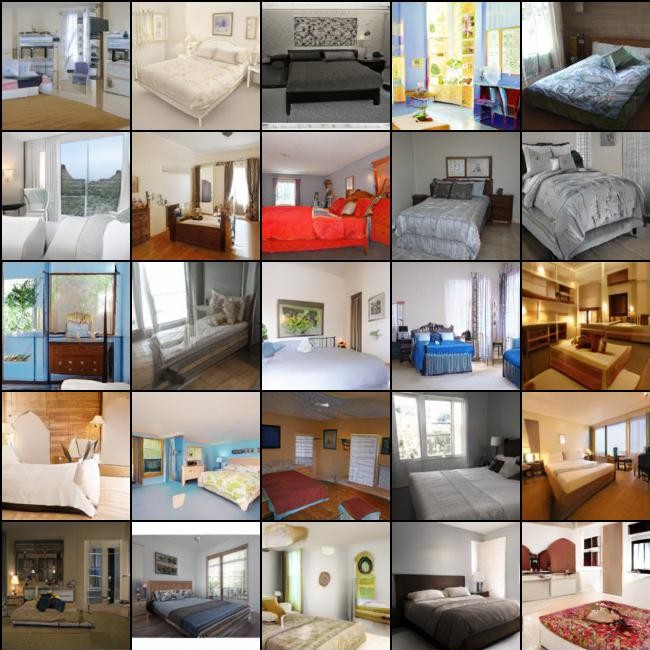}
  \caption{LSUN-Bedroom}
  \label{fig:6}
\end{subfigure}
\caption{Samples generated by our DISP-Vgg16 approach on large-scale image generation}
\label{app:large-scale-samples}
\end{minipage}
\end{figure*}

\end{document}